%% file: main.tex
\documentclass{article} 
\usepackage{iclr2027_conference,times}
\iclrfinalcopy

\input{math_commands.tex}

\usepackage{hyperref}
\usepackage{url}

\usepackage{graphicx}
\usepackage{booktabs}
\usepackage{subcaption}

\usepackage{tikz}

\title{Log-Depth Recurrent Language Modeling}

\author{%
Yiqin Wang \quad Nuri Cingillioglu \quad Charles Pert\\
Department of Computing\\
Imperial College London\\
London, United Kingdom\\
\texttt{\{kevin.wang23,nuric,charles.pert\}@imperial.ac.uk}
}

\begin{document}
\maketitle

\ificlrfinal
  \lhead{\small Preprint.}
\fi
\begin{abstract}
Language modeling using Transformers has become commonplace despite their fixed computational depth and quadratic runtime with respect to input tokens. Recurrent models on the other hand offer linear depth but no parallel execution. In this work, we extend balanced-tree recursive operators from sequence encoding to autoregressive prediction, enabling all prefix representations to be computed with logarithmic depth and linear runtime. Our experiments provide an initial characterization of this model class, demonstrating robust length extrapolation and performance approaching that of ALiBi-based Transformers, highlighting its potential as an alternative architecture for language modeling.
\end{abstract}

\section{Introduction}

The success of large language models (LLMs) demonstrates the power of learning from large-scale corpora for tasks including text generation, translation, summarization, and question answering~\citep{neto_automatic_2002,sutskever_sequence_2014,kalchbrenner-blunsom-2013-recurrent,brown2022language}. Much of this progress has been enabled by the Transformer~\citep{vaswani_attention_2023}, whose parallelism makes large-scale autoregressive training practical. During training, a Transformer computes $\{p(x_i\mid x_{<i})\}_{i=1}^{N}$ in parallel, but its computational depth remains independent of sequence length. Instead, the depth of recurrent models depends on sequence length~\citep{elman1990finding}, but conventional recurrence imposes a linear-depth sequential bottleneck. This motivates architectures that allocate more computation to longer sequences while retaining logarithmic depth and parallel computation.

Tree-structured recursive neural networks offer one direction toward this goal, but studies are typically limited to encoding tasks like sentiment analysis~\citep{tai_improved_2015,chen_sentence_2015,zhao_self-adaptive_2015,chowdhury2023recursion}. Applying these models to predict every autoregressive prefix would duplicate substantial computation, and developing efficient autoregressive variants remains an important open problem~\citep{ray_chowdhury_length_2024}. Log-Depth Recurrent Units~\cite[LDRUs;][]{pert_length_2026} apply associativity-biased operators in an up-sweep over a balanced tree, demonstrating generalization on regular-language encoding tasks. However, the up-sweep produces a single representation and does not provide an efficient way to compute the representation for every prefix.

In this work, we add the corresponding down-sweep to complete the Blelloch parallel scan~\citep{Blelloch1993Prefix}. By reusing the intermediate representations constructed during the up-sweep, the down-sweep produces representations for all sequence prefixes in parallel while retaining logarithmic depth and linear total work, as shown in Figure~\ref{fig:scan-diagram}. This augmentation yields a novel work-efficient parallel procedure for training these operators to predict the next token autoregressively.

We evaluate the resulting log-depth recursive models on three language modeling corpora of increasing sizes: Penn Treebank~\cite[PTB;][]{marcus_building_1993}, WikiText-2~\citep{merity_pointer_2016}, and OpenWebText2~\citep{pile}. Our aim is to provide an initial empirical characterization of where these models lie relative to conventional attention-based approaches, while identifying promising directions for their further development. Our contributions are threefold:
\begin{enumerate}
\setlength\itemsep{0cm}
\item We use the down-sweep from the scan to introduce a work-efficient parallel procedure for training log-depth operators autoregressively.
\item We provide an initial characterization of the performance of these operators on autoregressive language modeling across three corpora of different scales.
\item We evaluate the trained operators' ability to extrapolate to longer contexts.
\end{enumerate}

\section{Method}
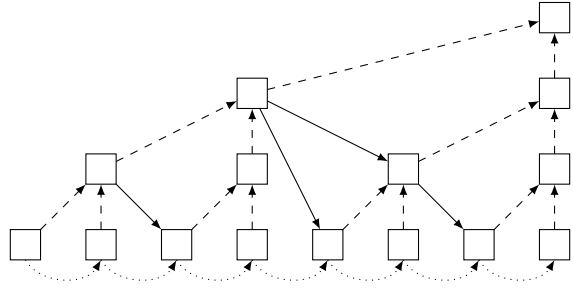
\begin{figure}[hbtp]
    \centering
    \begin{tikzpicture}[
        scale=1,
        square/.style={
            rectangle,
            draw=black,
            fill=white,
            minimum size=0.4cm,
            inner sep=0pt,
            outer sep=0pt,
            anchor=center
        },
        arrow/.style={->, >=latex}
    ]

        \foreach \x in {0,...,7} {
            \node[square] (L1\x) at (\x,0) {};
        }

        \foreach \x in {1,3,5,7} {
            \node[square] (L2\x) at (\x,1.0) {};
        }

        \foreach \x in {3,7} {
            \node[square] (L3\x) at (\x,2.0) {};
        }

        \node[square] (L47) at (7,3.0) {};


        \foreach \source in {0,1} {
            \draw[arrow, dashed] (L1\source) -- (L21);
        }
        \foreach \source in {2,3} {
            \draw[arrow, dashed] (L1\source) -- (L23);
        }
        \foreach \source in {4,5} {
            \draw[arrow, dashed] (L1\source) -- (L25);
        }
        \foreach \source in {6,7} {
            \draw[arrow, dashed] (L1\source) -- (L27);
        }

        \draw[arrow, dashed] (L21) -- (L33);
        \draw[arrow, dashed] (L23) -- (L33);
        \draw[arrow, dashed] (L25) -- (L37);
        \draw[arrow, dashed] (L27) -- (L37);

        \draw[arrow, dashed] (L33) -- (L47);
        \draw[arrow, dashed] (L37) -- (L47);


        \draw[arrow] (L33) -- (L25);
        \draw[arrow] (L25) -- (L16);

        \draw[arrow] (L33) -- (L14);

        \draw[arrow] (L21) -- (L12);


        \draw[]
            (L10.south) edge[dotted,arrow,bend right=60] (L11.south);
        \draw[]
            (L11.south) edge[dotted,arrow,bend right=60] (L12.south);
        \draw[]
            (L12.south) edge[dotted,arrow,bend right=60] (L13.south);
        \draw[]
            (L13.south) edge[dotted,arrow,bend right=60] (L14.south);
        \draw[]
            (L14.south) edge[dotted,arrow,bend right=60] (L15.south);
        \draw[]
            (L15.south) edge[dotted,arrow,bend right=60] (L16.south);
        \draw[]
            (L16.south) edge[dotted,arrow,bend right=60] (L17.south);

    \end{tikzpicture}
    \caption{Up-sweep (dashed) and down-sweep (solid) operations on a length-8 sequence. Dotted arrows indicate next-token prediction targets. The up-sweep builds hierarchical representations, reused by the down-sweep to produce all exclusive prefix representations.}
    \label{fig:scan-diagram}
\end{figure}

The parallel scan algorithm consists of an up-sweep followed by a down-sweep, as illustrated in Figure~\ref{fig:scan-diagram}. Given a sequence $e_1,\ldots,e_N$ and a binary composition operator $\odot$, the up-sweep recursively combines adjacent representations along a balanced binary tree, storing the intermediate results. The down-sweep reuses these intermediates to produce an exclusive prefix representation $h_i$ at every $i$ position. Although the classical scan assumes an associative operator, we do not enforce associativity; consequently, $h_i$ depends on the balanced tree structure. When $\odot$ is associative, the produced representations are invariant to the tree structure.

The up-sweep and down-sweep yield a structure of $O(\log N)$ depth and $O(N)$ number of operations where $N$ is the number of input tokens. Thus, we can compute all prefix representations with logarithmic depth and a linear number of operator applications, avoiding the linear sequential depth of a conventional recurrence and the quadratic runtime of Transformers, see Figure~\ref{fig:WikiText2-inferences}.

In our approach, the input to the parallel scan is the hidden representations of the tokenized sequence and the operator is a learnable neural network, $\odot_\theta:\mathbb{R}^{D}\times\mathbb{R}^{D}\rightarrow\mathbb{R}^{D}$. Given token embeddings $x\in\mathbb{R}^{N\times E}$, we apply layer normalization, a linear projection, and a position-wise residual MLP to obtain hidden representations $h\in\mathbb{R}^{N\times D}$. After each composition, we apply layer normalization and a residual post-merge network. The scan produces $h_i$ at every sequence position, which is projected to the embedding dimension and decoded using a tied token-embedding matrix. The model is trained using the standard next-token cross-entropy loss on shifted targets.

In our experiments, we parametrize $\odot_\theta$ as a Gated Recursive Cell~\citep[GRC;][]{shen_ordered_2019}. The GRC produces a gated mixture of a pair of states $a, b\in\mathbb{R}^D$. The two states are concatenated $[a;b]$, and passed through a two‑layer MLP. The MLP's hidden width is $4D$ with a $4D$ vector output, which is divided into three gate vectors and one candidate: $[g_a; g_b; g_c; c]\in\mathbb{R}^{4D}$, with $g_a,g_b,g_c,c\in\mathbb{R}^{D}$. The gates receive a sigmoid activation, and the operator's final output is a gated sum followed by layer normalization:
\begin{gather*}
\hat g_a,\hat g_b,\hat g_c = \sigma(g_a),\sigma(g_b),\sigma(g_c),\\
\odot_\theta(a, b)=\text{LN}\left(\hat g_a\circ a+\hat g_b\circ b+\hat g_c\circ c\right),
\end{gather*} where $\circ$ represents element-wise multiplication. We refer to GRC operators trained autoregressively as \emph{AR-GRC}. An alternative choice for $\odot_\theta$ is the associativity-biased MLP-LDRU operator of~\citet{pert_length_2026}. In an ablation on WikiText-2 (see Appendix~\ref{app:operator_ablation}), AR-GRC outperforms AR-MLP-LDRU, indicating that the non-associative $\hat g_c\circ c$ improves expressive capacity.

\section{Experiments}

\begin{figure*}[tbhp]
  \centering
  \includegraphics[width=1\textwidth]{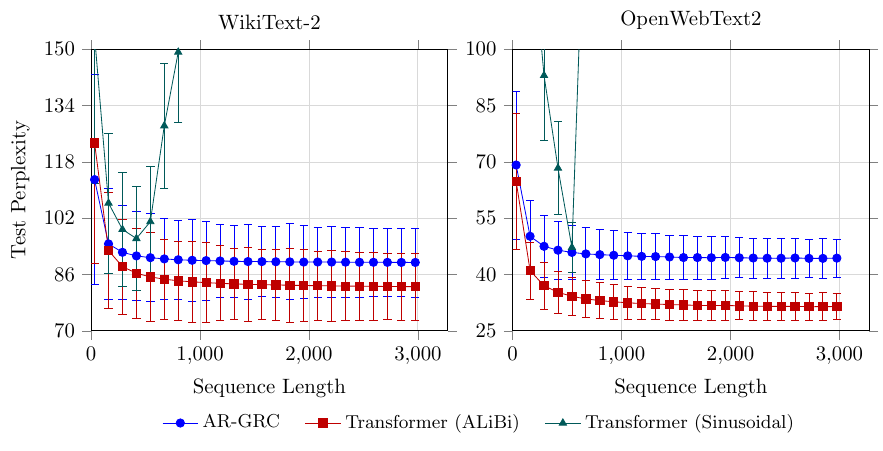}
  \caption{Perplexity as a function of sequence length on WikiText-2 and OpenWebText2. AR-GRC and Transformer (ALiBi) both indicate adaptability to long contexts; Transformer (Sinusoidal) explodes in perplexity at long contexts.}
  \label{fig:perplexity-context}
\end{figure*}

We train our models on datasets of increasing size: Penn Treebank~\cite[PTB;][]{marcus_building_1993}; WikiText-2~\citep{merity_pointer_2016}, and OpenWebText2~\citep{pile}. We use SentencePiece tokenizers with byte-pair encoding~\citep{kudo_sentencepiece_2018}, with vocabulary sizes of 8k for WikiText-2 and 50k for OpenWebText2. All models are trained on a maximum context length of 512 tokens. We compare against Transformer baselines with sinusoidal and ALiBi positional encodings~\citep{press2022train}, since these directly test whether AR-GRC preserves autoregressive prediction under context-length extrapolation. We match parameter counts of the models and aim for 20M parameters on WikiText-2 and 100M parameters on OpenWebText2. For PTB, AR-GRC has 3.5M parameters and the Transformers have 3.8M. We provide further experimental details in Appendix~\ref{app:experimental_setup}.

\textbf{Can AR-GRC models learn effective autoregressive next-token predictors across datasets and scales?} Table~\ref{tab:512_pplx_avg} reports test perplexity for each model across the three corpora. On PTB and WikiText-2, AR-GRC is broadly competitive with the Transformer baselines. The test perplexity of AR-GRC is 7.5--13.1\% higher on PTB, while on WikiText-2 it is 6.1\% higher than the ALiBi Transformer but 3.8\% lower than the sinusoidal Transformer. The gap is larger on OpenWebText2, where AR-GRC obtains 23.5--33.8\% higher perplexity than the Transformer baselines. These results demonstrate that the scan-based architecture can learn comparable autoregressive next-token predictors across datasets and scales, although its relative performance declined on OpenWebText2, a larger and more heterogeneous corpus.

\begin{table}[h!]
\centering
\caption{Test perplexity on PTB, WikiText-2 (WT-2) and OpenWebText2 (OWT2) for AR-GRC, Transformer (ALiBi) and Transformer (Sinusoidal). OpenWebText2 results are reported as mean $\pm$ standard deviation over three random seeds; PTB and WikiText-2 results are from a single seed. Lower is better.}
\label{tab:512_pplx_avg}
\begin{tabular}{lccc}
\toprule
\textbf{Model} & \textbf{PTB} & \textbf{WT-2} & \textbf{OWT2}\\
\midrule
\text{AR-GRC} & 14.34 & 90.74 & $46.08 \pm 0.06$ \\
\text{TF (ALiBi)} & 13.34 & 85.49 & $34.43 \pm 0.03$ \\
\text{TF (Sinusoidal)} & 12.68 & 94.30 & $37.32 \pm 0.10$ \\
\bottomrule
\end{tabular}
\end{table}

\textbf{Do AR-GRC models generalize to context lengths beyond those observed during training?}
We evaluate next-token prediction on test sequences ranging from 32 to 2,976 tokens in increments of 128, substantially exceeding the context length of 512 tokens. Figure~\ref{fig:perplexity-context} reports average test perplexity at each context length for WikiText-2 and OpenWebText2. On both corpora, AR-GRC maintains stable perplexity beyond the training-length boundary, exhibiting behavior similar to the ALiBi Transformer. By contrast, the perplexity of the sinusoidal Transformer increases sharply once the context length exceeds 512 tokens. These results show that the tree-structured computation learned by AR-GRC can be applied to sequences nearly 6 times longer than those observed during training without a corresponding degradation in next-token prediction performance.

AR-GRC does not rely on explicit positional encodings, which could make length extrapolation easier. The contrast between the two Transformer baselines demonstrates just how important the choice of positional encoding is for controlling extrapolation. While there has been recent work on Transformers without positional encodings~\citep{nope_transformer_still_pe}, their performance on language modeling is below the typical positional-encoding-based Transformers used here. Stable perplexity beyond training length demonstrates robustness to longer inputs, indicating AR-GRC is inductively biased toward autoregressive length extrapolation.

\begin{figure}[!htbp]
    \centering
    \includegraphics[width=0.5\linewidth]{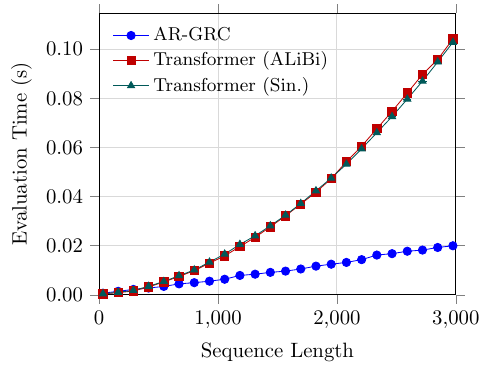}
    \caption{Model per-batch evaluation time as a function of sequence length on WikiText-2.}
    \label{fig:WikiText2-inferences}
\end{figure}

\textbf{How does AR-GRC evaluation time scale with context length compared with Transformer baselines?} During evaluation, we record the wall-clock time required for batched forward passes at each context length. Figure~\ref{fig:WikiText2-inferences} shows how runtime increases with sequence length for all models, but the observed scaling differs: both Transformer baselines grow quadratically, whereas AR-GRC grows linearly. The difference becomes increasingly pronounced at longer contexts, with AR-GRC running about $5\times$ faster than the Transformers at the maximum evaluated length for WikiText-2. The trends are the same on OpenWebText2, see Appendix~\ref{app:add_plots}.

This scaling behavior is consistent with the structure of the models. Attention computes a pairwise grid to model interactions between sequence positions, while AR-GRC applies $\odot_\theta$ a linear number of times in the length of the sequence. These measurements suggest that AR-GRC becomes increasingly advantageous as context lengths grow, despite not consistently outperforming the Transformer baselines. Further speed improvements may be possible by leveraging existing research in parallel scan implementations, such as~\citet{merrill2016singlepass}.

\section{Discussion}

These results provide the first evidence that balanced-tree recursive operators can be trained as practical autoregressive next-token predictors using a work-efficient parallel procedure. AR-GRC achieves perplexities close to established Transformer architectures on PTB and WikiText-2 and continues to learn meaningfully on the larger OpenWebText2 corpus. We therefore interpret these experiments as establishing the viability of this model class rather than identifying a fully optimized alternative to the Transformer. The performance gap motivates further study of the composition operator, model capacity, and mechanisms for selectively retaining relevant contextual information.

The fixed tree structure used by AR-GRC could be a limitation preventing better performance. Information is repeatedly compressed through predetermined composition paths, whereas self-attention allows each position to construct a token-dependent combination of the preceding context. This flexibility may become more important as the complexity and variability of the prediction task increase. However, the present experiments do not isolate the cause of the gap. It could be that the OpenWebText2 result is a limitation of the current AR-GRC instantiation and not evidence of a fundamental limitation of log-depth recursive models.

\section{Conclusion}
We adapt log-depth binary operators to autoregressive next-token prediction by incorporating the down-sweep of the parallel scan, enabling work-efficient training with AR-GRC. Across the three corpora, AR-GRC learns effective autoregressive predictors and exhibits robust length extrapolation comparable to the ALiBi Transformer.

Our experiments explore only a small part of the possible design space. More expressive binary operators may narrow the remaining performance gap, while larger-scale experiments are needed to determine how AR-GRC responds to increasing model size and training data. Establishing whether these architectures exhibit scaling laws similar to those of Transformers is left for future work. Another avenue for future study is to determine the associativity of the trained GRC operator, since non-associative composition makes representations dependent on the balanced tree structure.

These results establish a viable alternative for autoregressive language modeling and motivate further study of this comparatively underexplored model class.



\bibliography{refs}
\bibliographystyle{iclr2027_conference}

\appendix

\section{Appendix}
\label{sec:appendix}

\input{appendix}

\end{document}

%% file: math_commands.tex
\usepackage{amsmath,amsfonts,bm}

\def\eqref#1{equation~\ref{#1}}

\def\1{\bm{1}}

\DeclareMathAlphabet{\mathsfit}{\encodingdefault}{\sfdefault}{m}{sl}
\SetMathAlphabet{\mathsfit}{bold}{\encodingdefault}{\sfdefault}{bx}{n}



%% file: appendix.tex
\subsection{Binary Operator Ablation} \label{app:operator_ablation}
In this work, we focus on reporting performance achieved with GRC. This choice was informed by experiments on PTB and WikiText-2 varying $\odot_\theta$ between GRC and the MLP operator proposed by \citet{pert_length_2026}. We present these results in Table~\ref{tab:ablation_pplx}, showing that GRC outperforms the MLP operator.

\begin{table}[!htbp]
\centering
\small
\caption{Test perplexity on PTB and WikiText-2 for AR-GRC and AR-MLP-LDRU. Lower is better.}
\label{tab:ablation_pplx}
\begin{tabular}{lcc}
\toprule
\textbf{Model} & \textbf{PTB} & \textbf{WT-2} \\
\midrule
AR-GRC & 14.34 & 90.74  \\
AR-MLP-LDRU & 15.94 & 122.81 \\
\bottomrule
\end{tabular}
\end{table}

\subsection{Experimental Setup}
\label{app:experimental_setup}

We use the AdamW optimizer~\citep{adamw_optimizer}. The learning rate follows a linear warm-up phase followed by scheduled cosine annealing. We use early stopping with a patience of 10 epochs based on validation perplexity. We use a batch size of 32. There are no warm-up steps for PTB, 300 for WikiText-2, and 2,000 for OpenWebText2. We found that learning-rate warm-up improved performance. The vocabulary sizes are 1.5k for PTB, 8k for WikiText-2, and 50k for OpenWebText2.

Model hyperparameters are reported in Table~\ref{tab:hyperparams_all}. These were selected with a small Optuna~\citep{akiba2019optuna} sweep followed by light manual tuning for each dataset. For the OpenWebText2 Transformer, we used hyperparameters similar to those used by \citet{chi_transformer_2023}.

\begin{table}[htbp]
\centering
\setlength{\columnsep}{0pt}
\caption{Model hyperparameters for WikiText-2 and OpenWebText2.}
\label{tab:hyperparams_all}
\begin{tabular}{lcccc}
\toprule
& \multicolumn{2}{c}{\textbf{WikiText-2}} & \multicolumn{2}{c}{\textbf{OpenWebText2}} \\
\cmidrule(lr){2-3} \cmidrule(lr){4-5}
\textbf{Hyperparameter} & \textbf{AR-GRC} & \textbf{Transformer} & \textbf{AR-GRC} & \textbf{Transformer} \\
\midrule
Embedding Dimension    & 512  & 512  & 768   & 768  \\
Number of Layers       & 1    & 3    & 1     & 8    \\
Hidden Dimension       & 512  & 512  & 1536  & 512  \\
Number of Heads        & --   & 16   & --    & 12   \\
Learning Rate          & 1e-4 & 1e-4 & 5e-4  & 2e-4 \\
$L_2$ Regularization      & 1e-5 & 5e-5 & 0.0   & 1e-5 \\
Dropout Rate           & 0.1  & 0.3  & 0.0   & 0.15 \\
Post-embedding MLP   & True & -- & False & -- \\
\midrule
Parameter Count        & 16.6M & 17.6M & 97.4M & 95.1M \\
\bottomrule
\end{tabular}
\end{table}

\clearpage
\subsection{Additional Figures}
\label{app:add_plots}

\begin{figure}[htb]
\centering
\includegraphics[width=0.6\linewidth]{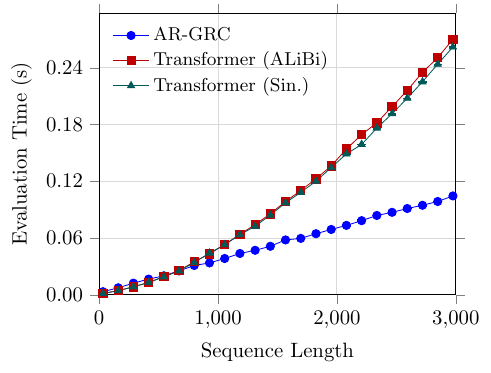}
\caption{Model per-batch evaluation time as a function of sequence length on OpenWebText2.}
\label{fig:ood_owt2}
\end{figure}